\documentclass[a4paper,conference]{IEEEtran}
\newcommand{\etal}{\textit{et al.}}

\usepackage[noadjust]{cite}

\usepackage{float} 
\usepackage[pdftex]{graphicx}
\ifCLASSINFOpdf
\else
\fi
\usepackage{color}
\usepackage{soul}
\def \b{\boldsymbol}
\def \RR{\mathbb{R}}
\def \SO{\mathrm{S\hspace{-1pt}O\hspace{-1pt}(3)}}
\def \half {\dfrac{1}{2}}

\usepackage{amsmath,amssymb}
\begin{document}
%
\title{Hybrid Approach for 3D Head Reconstruction: Using Neural Networks and Visual Geometry}



%


\author{\IEEEauthorblockN{Oussema Bouafif\IEEEauthorrefmark{1}\IEEEauthorrefmark{2},
Bogdan Khomutenko\IEEEauthorrefmark{1} and
Mohamed Daoudi\IEEEauthorrefmark{2}}
\IEEEauthorblockA{\IEEEauthorrefmark{1}MCQ-Scan, Lille, France}
\IEEEauthorblockA{\IEEEauthorrefmark{2}IMT Lille Douai, Univ. Lille, CNRS UMR 9189 CRIStAL, Lille, France}
\IEEEauthorblockA{Email: o.bouafif@mcq-scan.com, b.khomutenko@mcq-scan.com, mohamed.daoudi@imt-lille-douai.fr}}


\maketitle

\begin{abstract}
Recovering the 3D geometric structure of a face from a single input image is a challenging active research area in computer vision.
In this paper, we present a novel method for reconstructing 3D heads from a single or multiple image(s) using a hybrid approach based on deep learning and geometric techniques.
We propose an encoder-decoder network based on the U-net architecture and trained on synthetic data only.
It predicts both pixel-wise normal vectors and landmarks maps from a single input photo.
Landmarks are used for the pose computation and the initialization of the optimization problem,
which, in turn, reconstructs the 3D head geometry by using a parametric morphable model and normal vector fields.
State-of-the-art results are achieved through qualitative and quantitative evaluation tests on both single and multi-view settings. 
Despite the fact that the model was trained only on synthetic data, it successfully recovers 3D geometry  and precise poses for real-world images.
\end{abstract}


%

\section{Introduction}
In the last few years, several computer vision applications have used 3D face models. 
Three-dimensional face model reconstruction encounters different challenges, such as difficult orientations and illumination variations~\cite{DBLP:journals/prl/AbateNRS07}.
Several applications such as facial recognition~\cite{DriraPami2013}, facial animation~\cite{thies2016face2face}, gender classification~\cite{XiaPatternR2015}, and facial expression recognition~\cite{BenAmorIEEETC2014} have achieved better results using 3D face models. There is a number of different techniques to tackle the 3D face reconstruction problem.
Some approaches are based on structure from motion, optical flow, or shape from shading~\cite{kemelmacher20103d}.
However, these approaches all have limitations since most of them are sensitive to lighting conditions, reflections, shadows, and image quality.

Recently, machine learning has been used to solve these problems~\cite{richardson2017learning,dou2017end}. 
Neural networks are good at implicit modeling of complex variations of lighting, shading, and so on, which leads to a robust geometry estimation.
On the other hand, one of the most well-known difficulties in applying neural networks is the lack of 3D face data sets.
In some cases, this problem makes end-to-end learning systems less efficient than geometric methods.
The use of synthetic data or fitted 3D models from~\cite{kemelmacher20103d} has been proposed in many approaches to meet this need, even if the latter produces only an approximation of the ground truth, which may lead to a low reconstruction accuracy. 
Another drawback of end-to-end machine-learning-based approaches is the lack of control over the reconstruction process. 

Another challenge in 3D face reconstruction is geometry representation.
One possible way is a 3D Morphable Model~\cite{blanz1999morphable} (3DMM).
Such models guarantee that the result will be a plausible head reconstruction, and at the same time they are flexible and expressive enough to represent a wide variety of face morphologies. 
The challenge is to estimate the model parameters.
Several approaches to fit morphable models have been proposed \cite{sela2017unrestricted,thies2016face2face,guo2018cnn}. They are usually based either on a set of discrete landmarks \cite{yanga2018landmark} or on contours \cite{bas2016fitting}. 
We argue that pixel-wise fitting (used in some other 3D reconstruction applications \cite{engel2014lsd}) is the best way to exploit as much information present in the image as possible. 
But the application of this technique to 3D head reconstruction is problematic.
In the classical photometric approach, we would require texture and lighting model, which are generally not available. 

\begin{figure*}[!htbp]
 \centering
{\includegraphics[width=10.5cm,keepaspectratio]{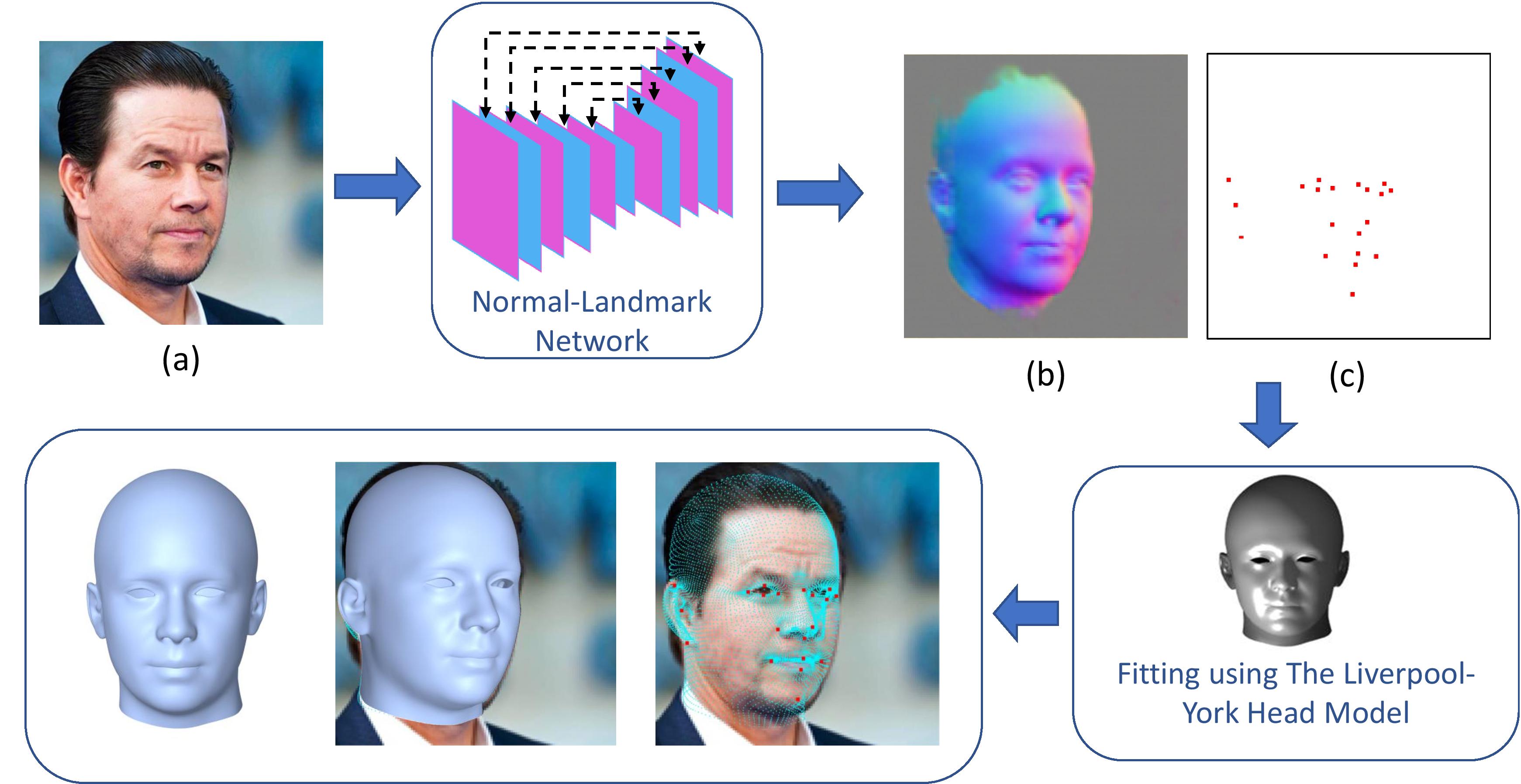}}
\caption{Overview of our proposed 3D face reconstruction method. Given an input facial image (a), we estimate two different maps (normal surface map $\cal N$ (b), Landmarks map $\cal Z$ (c)) used to reconstruct the 3D facial shape via a fitting process with the LYHM~\cite{Dai_2017_ICCV} morphable model.}
\label{fig:Overview}
\end{figure*}

In this paper, we propose a hybrid approach composed of deep learning and geometric optimization methods (See Fig.~\ref{fig:Overview}), capable of reconstructing a 3D head model from one or multiple facial images.
The key concept is that surface normals have a high degree of invariance (scale, translation), they don't depend on texture, shadows, and lighting; and they convey rich information about geometry. 
First, we use an encoder-decoder network that translates a facial input image into a landmarks map ($\cal Z$) and a facial normal map ($\cal N$).
Then, using these maps in a parametric regression algorithm, we reconstruct the 3D facial model.

The main contributions of this paper are: 
\begin{itemize}


\item The proposed approach predicts both pixel-wise normal vectors and landmarks maps from a single input photo.

\item We present a mesh fitting strategy based on surface normal vectors which address the reconstruction of the head directly from a single or multiple images and aims to recover a complete craniofacial human head as well as its pose.

\item We show that our approach achieves state-of-the-art performance on the BU-3DFE~\cite{yin20063d} data set. We show that the proposed model generalizes well to real-world images, even though it has been trained only on synthetic data.

\end{itemize}

\section{Related Work}

\subsection{Morphable Models}
The use of 3D morphable models (3DMM) is one of the fundamental approaches proposed to address 3D face reconstruction from a single input image. The most well-known model is perhaps the BFM model~\cite{blanz1999morphable}. This model is constructed using Principal Component Analysis (PCA) given a large set of 3D faces. 
By changing the model parameters, one can obtain a variety of 3D facial geometries. 

\subsection{Deep Learning Face Reconstruction Methods}

In the last few years, Convolutional Neural Networks (CNNs) have demonstrated better performances with fitting the 3DMM to an input image. 
For example, fitting via inverse rendering as it was shown in~\cite{thies2016face2face,guo2018cnn}, or regressing the 3DMM shape parameters directly from an input image~\cite{tuan2017regressing}.

Other techniques directly map the input photo onto diverse kinds of images which contain information on the geometry, such as~\cite{sela2017unrestricted}, which produces a depth map and a 3D correspondence map.
After that, a template mesh is deformed in order to fit these produced maps.
In~\cite{richardson2017learning} Richardson~\etal~proposed CoarseNet and FineNet to provide both coarse and fine details of facial shape.
Another method was proposed by~\cite{feng20183d}, which learns 3D face curves from horizontal and vertical epipolar plane images (EPI) of light field images utilizing a densely connected network (FaceLFnet).
This model-free approach produced curves that are combined together to obtain a more accurate combined 3D point cloud.
An end-to-end 3D face reconstruction method has been proposed in~\cite{dou2017end}, where the problem was divided into two sub-tasks: predicting the neutral 3D face and the expression parameters.
In~\cite{wu2019mvf} Wu~\etal~proposed a multi-view facial approach based on an end-to-end trainable CNN to regress 3DMM parameters via a differentiable dense optical flow estimator for the alignment and the photometric losses. In~\cite{feng2018joint}, a transfer function between an input image and the UV map was learned using a CNN. This UV map is a 2D representation of the 3D shape of the 3DMM. A direct approach was proposed in~\cite{jackson2017large} to reconstruct the 3D face with the use of a volumetric CNN regression.

To be able to use an unlimited amount of data for training, unsupervised learning methods were proposed.
A differentiable renderer was used in~\cite{genova2018unsupervised} for the optimization task after learning 3DMM parameters from facial identity encoding.
In~\cite{sanyal2019learning}, RingNet was trained using a shape consistency loss similar to the triplet loss to regress all images of the same person to a unique latent form vector. A novel deep 3D face reconstruction approach in~\cite{deng2019accurate} uses a hybrid loss function for weakly-supervised learning.

More recently, Wang~\etal~\cite{wang2019digital} uses an off-the-shelf face recognition neural network trained on high-quality synthetic data to fully reconstruct facial geometries based on a single selfie.
Inspired from~\cite{trigeorgis2017normals} and based on a GAN model, Face-Normal-Net~\cite{bouafif2020monocular} learned from a synthetic data set to regress a normal map, used in a weighted least square solver to reconstruct a surface approximating the 3D shape. 
The DF2Net network~\cite{zeng2019df2net} uses three combined modules, which are each trained on a separate type of data set with different training strategies. 
The model progressively refines the subtle facial details like small crow’s feet and wrinkles.
MMFace~\cite{yi2019mmface} is a multi-metric regression network composed of two sub-networks: a volumetric sub-network to estimate an intermediate face geometry and a parametric sub-network to infer the corresponding 3DMM parameters.

\subsection{Normal Map Prediction}

Multiple works in computer vision have been proposed to solve tasks such as depth map or surface curvature prediction, semantic segmentation, and edge detection.
Among these works, prediction of the normal map has been studied by different learning-based approaches since producing high-quality normal maps for complex objects represents a challenging task.
Several methods have been dedicated to the generation of normal maps from (outdoor/indoor) scenes or objects.
In~\cite{bansal2016marr}, a skip-network model has been trained to predict surface normal which is the input to another two networks that estimate the pose and the style of the object image.
In~\cite{qiu2019deeplidar}, $DeepLiDAR$ has been proposed to predict depth map with a normal surface of outdoor scenes from sparse LiDAR Data and a gray-scale image.

Another category of normal-map-prediction methods has been proposed, where the network takes sketches as input images.
In~\cite{hudon2018deep}, Hudon~\etal propose a multi-scale representation of the input images to ensure accuracy of produced maps.
An interactive method was proposed by Su~\etal~\cite{su2018interactive} where they proposed a conditional GAN framework based on U-Net model~\cite{ronneberger2015u}.
As we present here, another type of approach has proposed to predict normal maps from facial images to use it in a 3D reconstruction task.
In~\cite{trigeorgis2017normals}, a fully convolutional network trained on various synthetic and real-world data sets.
The obtained normal map is then used in the Frankot-Chellappa method~\cite{DBLP:journals/pami/FramkotC88} to reconstruct 3D facial shape. In~\cite{sengupta2018sfsnet}, a novel architecture SfSNet learns from a mixture of labeled synthetic and
unlabeled real images to solve the problem of inverse face rendering. It utilizes residual blocks to disentangle normal and albedo maps into separate subspaces.
More recently, Bouafif~\etal~\cite{bouafif2020monocular} trained a GAN model to generate normal maps from synthetic facial images which are used in a variational 3D reconstruction method. In~\cite{abrevaya2020cross}, a new cross-modal learning architecture was proposed to resolve the limited available ground truth data. Core to this approach is a novel module called deactivable skip connections, which allows integrating both the auto-encoded and image-to-normal branches within the same architecture that can be trained end-to-end. This allows learning of a rich latent space that can accurately capture the normal information. 

\subsection{Training With Synthetic Data}

Several facial analysis approaches like facial recognition \cite{kortylewski2018training}, face detection \cite{han2018improving} and facial expression analysis \cite{abbasnejad2017using} have utilized synthetic data to overcome the lack of large-scale training data during the use of neural networks.
3D face reconstruction is among these applications.
In~\cite{tuan2017regressing,zhu2016face}, a semi-synthetic data set was produced by implementing an optimization-based algorithm with proven precision to a database containing real-world faces.
Other approaches~\cite{guo2018cnn} have also reinforced this track by integrating a technique of inverse rendering.
Another option was brought up by~\cite{richardson2017learning,dou2017end,sela2017unrestricted,zeng2019df2net,richardson20163d}, which consists in producing a full synthetic data set. 
For example, they use the BFM morphable model~\cite{blanz1999morphable} for identities with Face Warehouse~\cite{cao2013facewarehouse} or 3DDFA~\cite{zhu2015high} model for expressions and producing with it, several types of images according to the need of the method. 
However, the performance of these methods might be limited by lighting models used for rendering, as well as the representativity of the subjects participated in the acquisition to create the model.

To deal with some of these problems, a short time ago, novel approaches have been proposed to boost realism in 3D synthetic face data sets.
To achieve that, more productive models and more accurate rendering techniques are suggested.
\cite{feng20183d} proposed a photo-realistic light field image synthesis method to generate a large-scale Epipolar Plane Images (EPI) data set selecting some examples from the BU-3DFE~\cite{yin20063d}.
In~\cite{wang2019digital}, Wang~\etal proposed an augmented 3D head data set with UV texture and a high-quality engine for rendering.
More recently, the LYHM~\cite{Dai_2017_ICCV} morphable model was used in~\cite{bouafif2020monocular} with  hair model \cite{hu2015single} and texture to construct a synthetic data head generator that can be used in various approaches related to facial analysis.

\section{Proposed Method}

In this section, we describe the details of our proposed framework as shown in Fig.\ref{fig:Overview}.
Our 3D reconstruction method takes a facial image as the only input and produces two output maps from an encoder-decoder model.
These two maps are a normal surface map $\cal N$ and a map which contains facial landmarks $\cal Z$.
We use these outputs in a 3D fitting algorithm to recover the 3D Model of the head. 
In Section \ref{sec:data_generator}, we describe our synthetic 3D head generator which allowed us to produce our training data set. 
In Section \ref{sec:CNN}, we show our network architecture which regresses two different maps that are aligned with the input image. 
The first is a normal surface map ($\cal N$). 
The second output is a map that contains facial landmarks ($\cal Z$). 
Finally, we explain the details of 3DMM fitting in Section \ref{sec:problem_formulation}.

\begin{figure}[!htbp]
\centering
{\includegraphics[width=6cm,keepaspectratio]{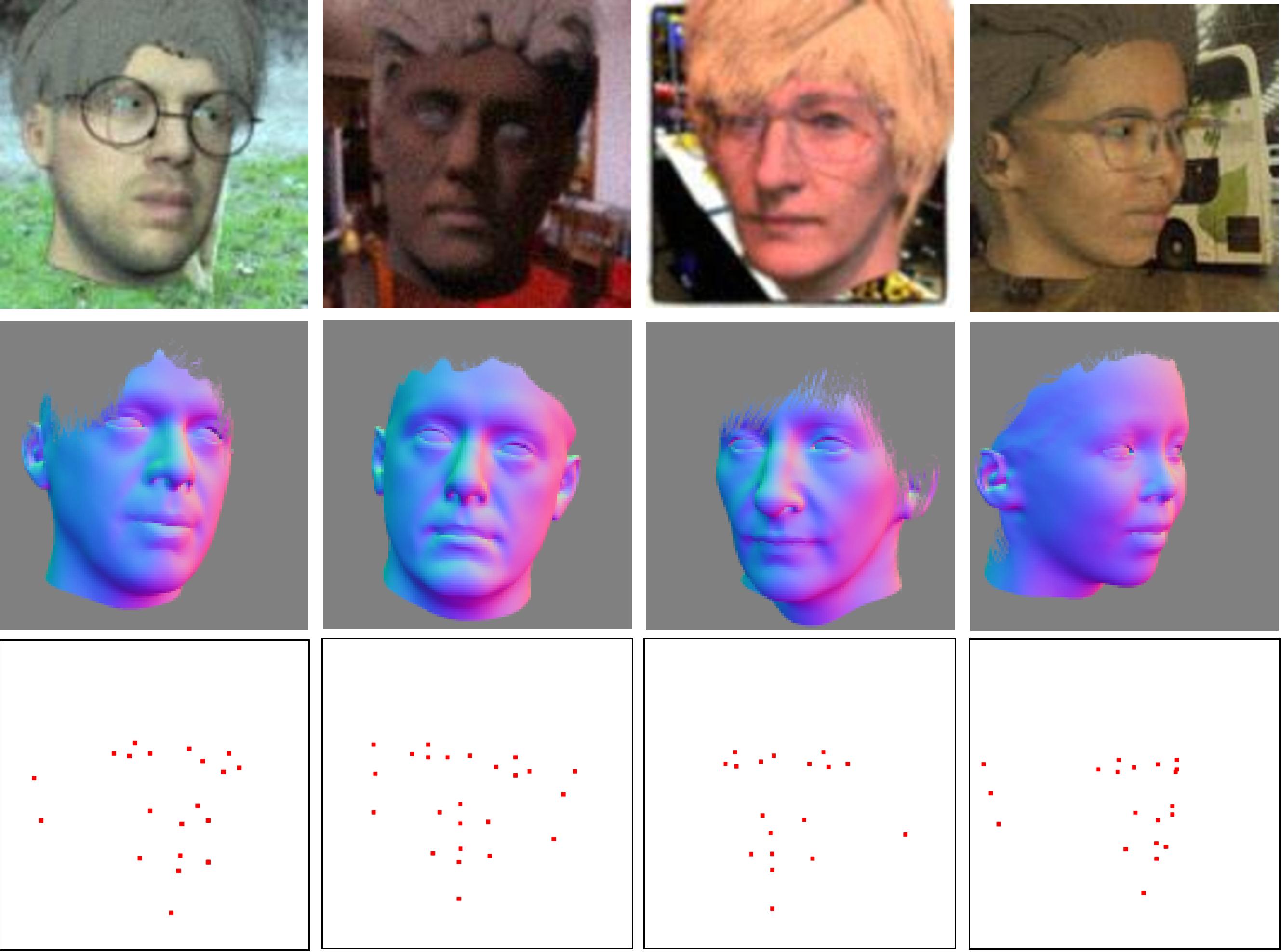}}
\caption{Training data samples. From top to bottom: Synthetic facial images, Normal surface maps $\cal N$ and Landmarks maps $\cal Z$.}
\label{fig:dataset_generation}
\end{figure}

\subsection{Synthetic Data Generation}
\label{sec:data_generator}

We propose a fully synthetic data set for deep neural network training. To achieve this, we have set up a synthetic data generator based on the one described in~\cite{bouafif2020monocular}.
The main source of information for geometric fitting is the normal map $\cal N$ produced by the face generator (second row in Fig.~\ref{fig:dataset_generation}).

To ensure alignment during the adjustment process, we have selected 24 vertices on the head model which are distributed over the eyes, nose, mouth, chin, and ears.
After that, the points that are not hidden by the hair or difficult poses are projected, and the landmark map ($\cal Z$) is generated (third row in Fig.~\ref{fig:dataset_generation}). This map is composed of 24 channels, each one containing a positive value at the corresponding landmark projection and zeros everywhere else.

\subsection{Network Architecture}
\label{sec:CNN}

\begin{figure}[!htbp]
 \centering
{\includegraphics[width=8.6cm,keepaspectratio]{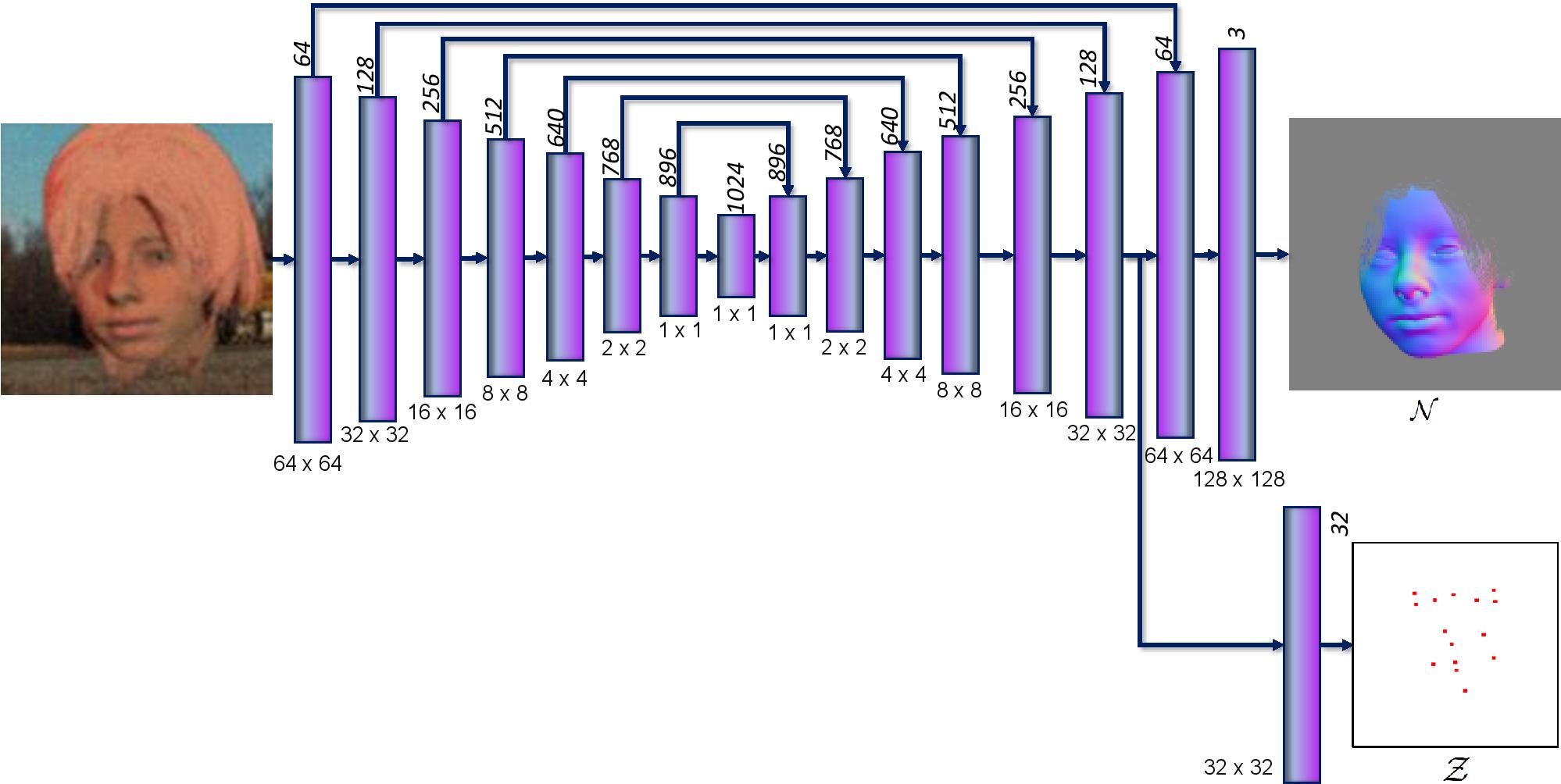}}
\caption{Details of Normal-Landmark Network. Our encoder-decoder architecture produces two different maps ($\cal N$ and $\cal Z$) (shown on the right) from a facial input image (shown on the left). The spatial size and number of layers are shown below and above each block, respectively.}
\label{fig:CNN-Diagramme}
\end{figure}

We propose for the network architecture an encoder-decoder based on the Su~\etal~\cite{su2018interactive} model which trained a GAN model to map a normal surface from a sketch and binary point mask inputs. 
This model uses the symmetric connection between the encoder and the decoder to reduce the information loss among successive layers, thus implementing the U-Net model~\cite{ronneberger2015u}.

In order to adapt the network to our problem, a certain number of changes have been made. 
The discriminator has been discarded.
For the training data, we give for the input the facial image (RGB), while we have only two generated maps $\cal N$ (3 channels) and $\cal Z$ (24 channels).
For the loss function, we use a composed pixel-wise loss to measure the difference between the generated maps and the real input ones as described here:
\begin{equation}
\begin{gathered}
\mathrm{Loss} =  L_{\cal N} + L_{\cal Z} \\
L_{\cal N} = ||{\cal N}_{\textrm{GT}} - {\cal N}||_2^2 \quad,\quad L_{\cal Z} = ||{\cal Z}_{\textrm{GT}} - {\cal Z}||_2^2
\end{gathered}
\label{eq:cnn}
\end{equation}
Where it enforces the rendered images $\cal N$ and $\cal Z$ to be similar to the inputs $\cal N_{\textrm{GT}}$ and $\cal Z_{\textrm{GT}}$ respectively. The architecture of Normal-Landmark Network is
illustrated in Fig.~\ref{fig:CNN-Diagramme}.
\subsubsection{Training Details}
\label{exp:training_settings}
To train our encoder-decoder model, we generate 60,000 facial images scaled to $128$ x $128$ (equally distributed between males and females) and their corresponding $\cal N$ and $\cal Z$ maps.
We train the model for about $3000$ epochs with a learning rate of $1e-5$, $32$ as batch size, and we adopt RMSprop as the optimizer.
To make the training images more realistic we add random blur effect and Gaussian noise as data augmentation.
When minimizing energy function in \eqref{eq:fitting}, we use $\lambda_{\cal N} = 1$ , $\lambda_{\cal Z} = 0.8$ and, $\lambda_{P} = 0.4$.

\subsection{3D Morphable Model fitting}
The core idea of the present work is the geometry-based fitting of a morphable model to a predicted field of normals,
using a camera projection model. This subsection describes in details the problem formulation.
\label{sec:problem_formulation}
\subsubsection{3D Morphable Model}
In our reconstruction method, we use the LYHM~\cite{Dai_2017_ICCV} morphable model.
By adding parameterized deformations to the mean face model $X_0$, we can generate variability in head identity, as follows:
\begin{equation}
\begin{aligned}
X &= X_{0} + W\b y
\end{aligned}
\label{eq:lyhm}
\end{equation}
Where $X \in \RR^{3 \times N_{\!X}}$ is the generated 3D head with $N_{\!X} = 11,510$ vertices; $X_0 \in \RR^{3 \times N_{\!X}}$ is the mean face, computed over the aligned facial 3D scans in the Headspace \cite{duncan2017novel} collection;
$\b y \in \mathbb{R}^{N_{\!y}}$ is the identity parameter vector with $N_{\!y} = 100$;  $W \in \mathbb{R}^{3 \times N_{\!X} \times N_{\!y}}$ is a tensor representing principal components of the shape model.

\subsubsection{Morphable Model's Normals}
In order to fit the normal map $\cal N$, the normal vector $\b n_{i}$ is calculated for each vertex location $\b p_i \in \mathbb{R}^3$, given the set of adjacent vertices $\left\{ \b q_{i,1}, \b q_{i,2}, ..., \b q_{i,M} \right\} \subset \mathbb{R}^3$. First an unnormalized version $\Tilde{\b n}$ is computed:

\begin{equation}
\begin{gathered}
\Tilde{\b n}_{i} = \sum_{j=1}^{N_{\!X}} \frac{(\b q_{i,j} - \b p_{i}) \times (\b q_{i,j+1} - \b p_{i})}{\|(\b q_{i,j} - \b p_{i}) \times (\b q_{i,j+1} - \b p_{i})\|}
\end{gathered}
\label{eq:normal}
\end{equation}
Then the actual normal $\b n_i$ is obtained by a simple vector normalization:
\begin{equation}
\b n_i = \frac{\Tilde{\b n}_i}{||\Tilde{\b n}_i ||}
\label{eq:normal_normal}
\end{equation}
This is not the most exact method of computing normals, but it is fast, which is important since it is computed at every optimization iteration.

\subsubsection{Projection Model}
In order to get the reference normal for each vertex $\b p$,
the latter is projected onto the image plane of $\cal N$ using the Pinhole camera model. The bi-cubic interpolation is used to get the corresponding reference normal from $\cal N$ (which is a discrete grid).
The projection process can be expressed as follows:
\begin{equation}
\begin{gathered}
\b a = \frac{1}{z}K  R  ( \b p + \b t ) 
\end{gathered}
\label{eq:3DMM_proj}
\end{equation}

Where $\b a \in \RR^2$ is the projected point; $R \in \SO$ is the rotation matrix, parametrized by  the three rotation angles of roll, pitch, and yaw, and denoted by $\b r$; and $\b t \in \RR^3$ is a translation vector.
In this context $z$ in the denominator represents the normalization of a 3D point to bring it to the normal plane 
($z$ is taken after multiplying the vector by $R$ but before multiplying it by $K$).
The projection matrix K contains three parameters $f, u_0, v_0$ and defines the projection model:
\begin{equation}
K = \begin{pmatrix} 
f \;& 0 \;& u_0 \\
0 \;& f \;& v_0
\end{pmatrix}
\end{equation}
Since the camera calibration is approximative, we think that it is enough to have only three intrinsic camera parameters.

\subsubsection{Parametric Regression}
Our implementation follows standard practices
(for example, \cite{guo2018cnn,richardson2017learning,sela2017unrestricted,tuan2017regressing}) which use the morphable model in a fitting process.
It aims to form an image as close as possible to a target image by finding the most suitable combination of parameters.
The target images ($\cal N$ and $\cal Z$) are produced from our encoder-decoder network described in Section \ref{sec:CNN}.
The problem can be expressed as the minimization of an energy function that represents the error between the produced maps and those generated by the morphable model. 
Our energy function contains three main components $E_{\cal N}$, $E_{\cal Z}$ and $E_{P}$:

\begin{equation}
\begin{gathered}
E = \lambda_{\cal N} E_{\cal N} + \lambda_{\cal Z} E_{\cal Z} + \lambda_{P} E_{P} \\
\end{gathered}
\label{eq:fitting}
\end{equation}

Let us describe each term.
$E_{\cal N}$ represents the difference between interpolated normals $\cal N(\b a)$ from the 2D projection of the morphable model onto $\cal N$ and vertex normals $\b n$  which are computed from the Morphable model using \eqref{eq:normal}. $E_{\cal N}$ is defined by:

\begin{equation}
E_{\cal N} = \half\sum_{i = 1}^{N_{\!X}}||{\cal N}(\b a_i) - \b n_i||^2
\label{eq:EN}
\end{equation}

The landmarks loss $E_{\cal Z}$ is defined as the distance between the detected landmarks $\b z_j$ and the projections $\b b_j$ of the corresponding subset of 3D vertices using the same projection model described in Eq.\ref{eq:3DMM_proj}. 

\begin{equation}
\begin{gathered}
E_{\cal Z} = \half\sum_{j = 1}^{N_z} ||\b z_j - \b b_j||^2 
\end{gathered}
\label{eq:EZ}
\end{equation}

Finally, the shape prior loss $E_P$ ensures plausibility of reconstructed heads by assuming priors given by the statistical head model represented by singular values $\sigma_k$.
In most 3DMM reconstruction methods~\cite{blanz1999morphable,genova2018unsupervised,guo2018cnn,sela2017unrestricted,wang2019digital,wu2019mvf}, this term is used to prevent the degeneration of the reconstructed geometry:

\begin{equation}
\begin{gathered}
E_{P} = \half\b y ^T {\cal C}^{-1}_y \b y 
\end{gathered}
\label{eq:EP}
\end{equation} 
${\cal C}_y$ is the covariance matrix, which in this case is a diagonal matrix containing $\sigma_1^{2} \!,\sigma_2^{2} \!,.. \sigma_{N_{\!y}}^{2}$.

The goal of the parameter regressor is to predict morphable model parameters $y$ and those related to the projection of the model from 3D space to 2D one. This can be formulated as a minimization problem as the following:

\begin{equation}
{\b y}^*, {\b t}^* , {\b r}^* , {K}^* = \underset{{\b y}, \b r, \b t, K   }{\mathrm{argmin}}\;E
\end{equation}
where $\b y$ is the head shape parameter vector, $\b t$ and $\b r$ are the translation and rotation parameters defining the head pose with respect to the camera, $K$ is the intrinsic camera matrix. 
This is a non-linear least-square problem which can be solved efficiently using the Levenberg-Marquardt algorithm.
The Ceres solver \cite{ceres-solver} is used as the optimization back-end.

\subsubsection{Head Pre-Alignment}
If we use only normals to fit the head model, we may encounter a certain number of convergence issues. First, it might take some time for the optimizer to converge to the desired position and orientation. Second, The optimization might get stuck in a local minimum. It is a good thing for all local-search optimizers to have a good initial guess.
We can achieve it by computing a plausible initial pose. It is done by optimizing $E_{\cal Z}$ over $\b t_0, \b r_0$, while fixing $\b y$ to 0 and $K$ to a certain prior value, in other words, by using the mean head model and landmarks given by the network. 
Since it is a purely geometric optimization problem, it almost always converges to the global solution and does it quickly, considering the low number of errors (up to 48 in our case) and unknowns (6 DoF).
This approach allows us to make the system robust with respect to the exact head pose and orientation in the image.




\subsection{Multi-view 3D Face Reconstruction}

\begin{figure}[!htbp]
 \centering
{\includegraphics[width=8cm,keepaspectratio]{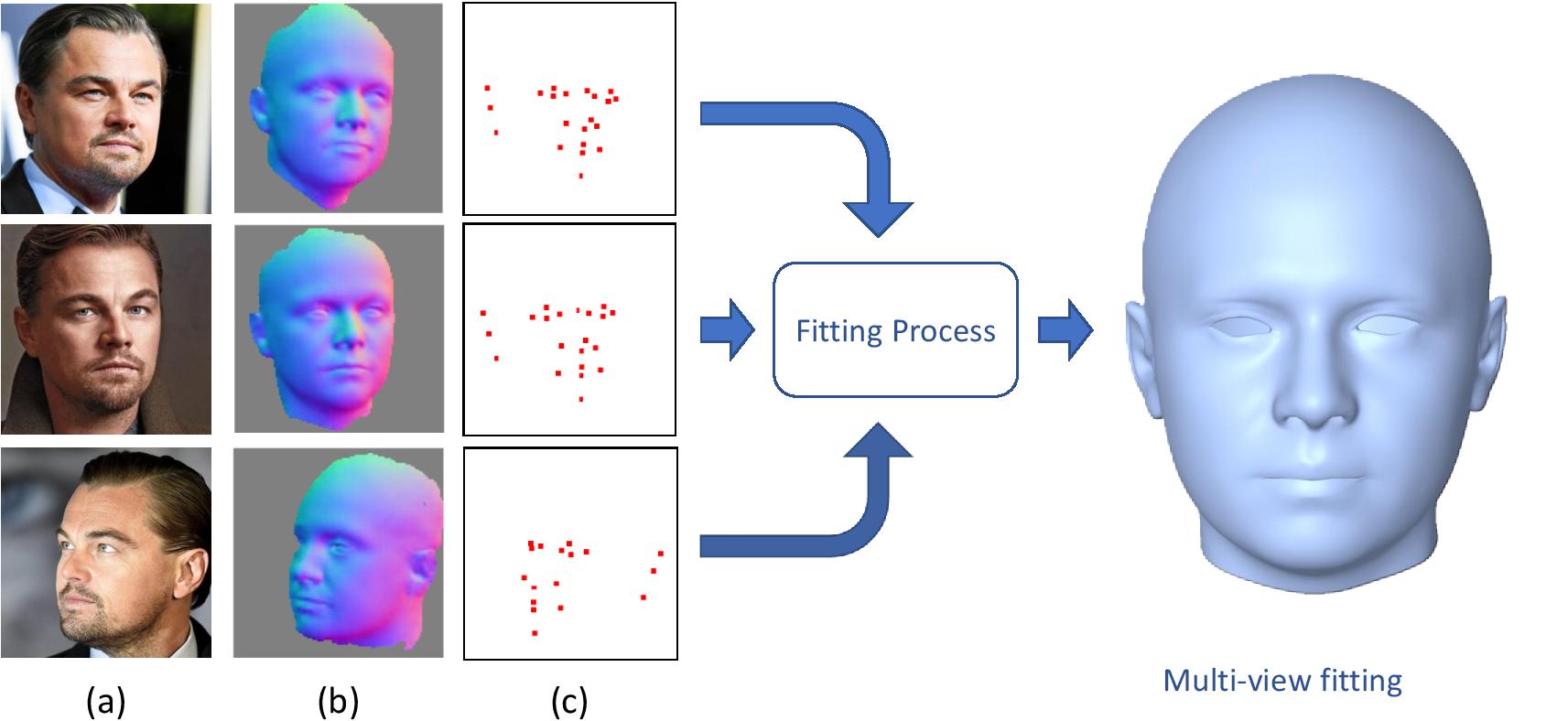}}
\caption{An illustration of our method for multi-view fitting. We estimate the two maps $\cal N$ (b) and $\cal Z$ (c) from each input image (a), then we use them in the same fitting process to get a unique 3D head reconstruction.}
\label{fig:Multi-image}
\end{figure}

Our approach can be employed to regress 3DMM parameters from multiple facial images of the same person in different views.
Note that our approach can be generalized to any number of input images. 
We produce ${\cal N}_{i}$ and ${\cal Z}_{i}$ maps for all input images, then we use output maps in the same fitting process.

Poses and camera parameters are computed independently for each image while $\b y$ is shared for all images.
In Fig.~\ref{fig:Multi-image}, we show an example of 3DMM fitting from multiple images of the same person.

\section{Experiments and Results}
In this section, we evaluate the quality of our 3D head reconstruction results obtained using the proposed framework.
Firstly, we show the qualitative results in Section \ref{exp:qualitative_evaluation}. 
Section \ref{exp:quantitaive_evaluation} is dedicated to the ability of the system to reconstruct 3D heads from real-world images by quantitative experiments on a benchmark database with available ground truth, where we compare our results to state-of-the-art methods.

\subsection{Qualitative Evaluation}
\label{exp:qualitative_evaluation}

\begin{figure*}
 \centering
{\includegraphics[width=10.5cm,keepaspectratio]{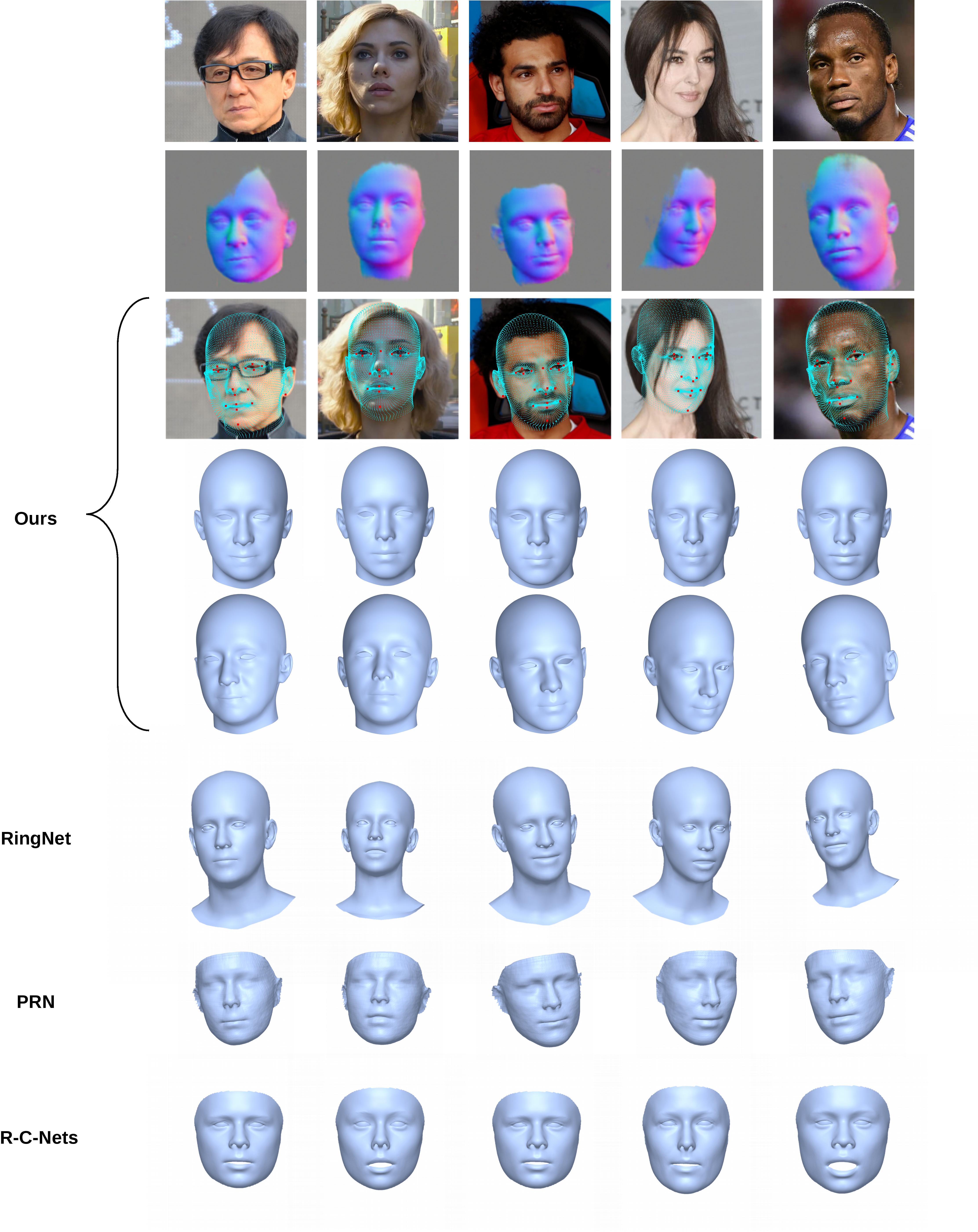}}
\caption{Examples of visual comparisons from some celebrities facial images with other methods. Rows contain in order; input image, predicted $\cal N$ map, input image with predicted landmarks (red squares) and dense alignment results (projected vertices of the 3DMM produced by our fitting process), Ours (frontal view), Ours (aligned) RingNet~\cite{sanyal2019learning}, PRN~\cite{feng2018joint} and R-C-Nets \cite{deng2019accurate}}
\label{fig:RealExamples}
\end{figure*}

In Fig.~\ref{fig:RealExamples}, we show our monocular 3D reconstruction process on the photos of certain celebrities captured from different poses.
This evaluation demonstrates how the network learns to represent the head structures of real people.
From row 2, we notice our network produces high-quality normal maps as well as accurate landmarks, although it was trained with fully synthetic data.
It successfully separates the head from the hair and the background.
One can see our method produces high-quality results, which fit the overall structure well.
As we use the LYHM~\cite{Dai_2017_ICCV} morphable model that includes the cranial part, our method allows us to recover the 3D model of the head.
The cranial region approximates the areas hidden by the hair whereas we do not estimate the invisible parts of the face skin.

\begin{figure}[!htbp]
 \centering
{\includegraphics[width=8.5cm,keepaspectratio]{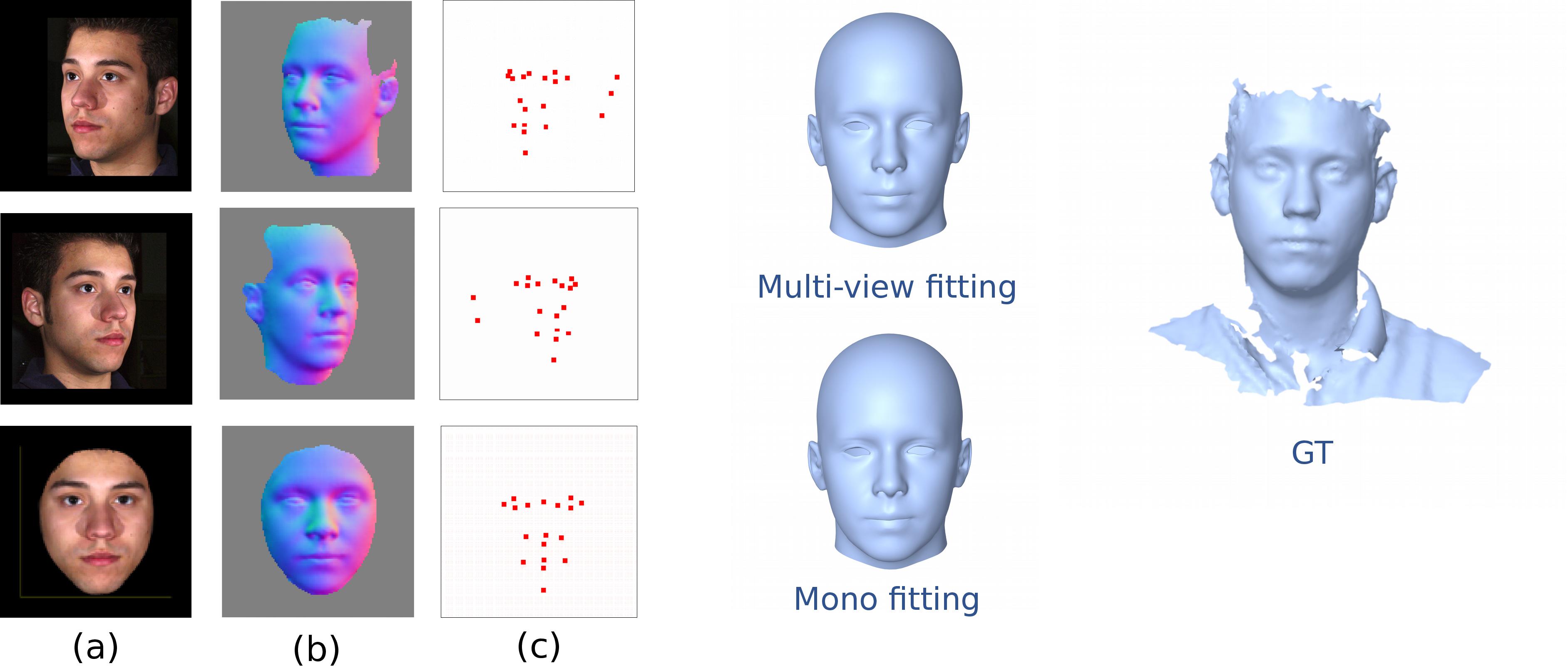}}
\caption{An illustration of stereo and mono fitting from BU-3DFE~\cite{yin20063d} data set example. Input images (a), normals surface map $\cal N$ (b), landmarks map $\cal Z$ (c). (Multi-view fitting): 3D head reconstruction using all images in the same fitting process. (Mono fitting) : 3D head fitted exploiting only the frontal image in the fitting process (third row). (GT) : the ground-truth of the 3D head mesh.
}
\label{fig:BU-3DFEEval}
\end{figure}

A comparison between the monocular and a multi-view processes using an example of the BU-3DFE~\cite{yin20063d} data set is illustrated in Fig.~\ref{fig:BU-3DFEEval}.
We can notice a significant visual resemblance between the 3D reconstruction and the ground truth using the multi-image process---the 3D reconstruction better captures the shape of the face. We notice that for some of the other examples in BU-3DFE, the results are less accurate.

\subsection{Quantitative Evaluation}
\label{exp:quantitaive_evaluation}


For quantitative evaluation, we demonstrate the effectiveness of our approach using the BU-3DFE~\cite{yin20063d} data set, which contains 100 subjects (56 female, 44 male), with different ages and a variety of ethnic/racial ancestries, and where each 3D model has corresponding 2D images captured under controlled settings.
In our evaluation, we use both frontal and profile images of all subjects to evaluate our algorithm's performance with both mono and multi-view reconstruction.
Since our 3DMM does not contain expressions, we use only neutral expressions faces.
We performed a rough pre-alignment process between the reconstructed model and the ground truth using 6 pre-selected vertices.
Next, the alignment and registration processes with the ground truth model are performed using the iterative closest point (ICP) solver, after that, we compute the point-to-plane distances and the absolute depth errors between the reconstructed 3D models and the ground-truth 3D meshes.
We eliminate examples when the alignment process fails (which is about 3-4\%).

We use the point-to-plane Root Mean Square Error (RMSE) as the performance measure. 
We argue that it is a good way of evaluating performance, which is not sensitive to either the number of vertices in the model or the way of how they are distributed. 
It is computed as follows. Once the models are aligned, for each vertex $\b p_i$ of the reference model, we compute the nearest neighbor $\b q_i$ from the reconstructed model to be evaluated, as well as normal vectors $\b n_i$ of the reference model. Second, we compute the error projected to the normal, and compute the RMSE:
\begin{equation}
\varepsilon = \sqrt{\frac{\sum_{i = 1}^N\left((\b p_i - \b q_i) \cdot \b n_i\right)^2}{N}}
\end{equation}
where $N$ is the number of vertices of the reference model. 
Once $\varepsilon$ is computed for each subject, we can find the mean and standard deviation across the benchmark data set.
\subsubsection{Results and Discussion}
In Table~\ref{tab:Evaluation1}, we report the numeric results in the form of $\mathrm{mean} \pm \mathrm{std}$ for the point-to-plane (RMSE) and for the absolute depth errors, we report mean ($\mu$), standard deviation ($\sigma$), median ($\widetilde{m}$), and the average ninety percent most significant error ($\delta_{90\%}$). 
We have compared our system in both monocular (Mono) and multi-view (Multi) settings to the state-of-the-art systems with code available online \cite{sanyal2019learning,feng2018joint,deng2019accurate}. We used the exact same procedure for all the tested systems to make sure that all of them are in the same conditions.
We can see that the mean error of the multi-view approach is lower than that which we have with the monocular configuration. In most cases, the morphology of the entire head is better captured using several images. In some other cases, this multi-view approach gives less good result. And since the evaluation is only done on the facial part, it does not contribute to the precision.

One can see that the proposed method is performing at the level of the state of the art, being clearly outperformed only by the R-C-Nets. The precision obtained by us for the latter is somehow worse than the one reported in the corresponding publication \cite{deng2019accurate}, which is $1.40\pm0.31$ for the point-to-plane (RMSE) and which can be due to some differences in the evaluation technique.

\setlength{\tabcolsep}{1pt}
\begin{table}[!h]
\begin{center}
\caption{Quantitative comparison on the BU-3DFE~\cite{yin20063d} data set. Lowers are better}
\label{tab:Evaluation1}

\begin{tabular}{|c|c|c|c|c|c|}
\hline
Method & Ours (Mono) & Ours (Multi) & RingNet \cite{sanyal2019learning}  & PRN \cite{feng2018joint} & R-C-Nets \cite{deng2019accurate}\\
\hline
RMSE & 1.74 $\pm$ 0.44 & 1.67 $\pm$ 0.43 & 1.90 $\pm$ 0.49 & 1.86 $\pm$ 0.47 & \textbf{1.60 $\pm$ 0.41} \\
\hline
$\mu$ & 2.21 & 2.17 & 3.42 & 1.83 & \textbf{1.64} \\
\hline
$\sigma$ & 1.08 & \textbf{1.04} & 1.58 & 1.70 & 1.69 \\
\hline
$\widetilde{m}$ & 2.09 & 2.05 & 3.23 & 1.43 & \textbf{1.27} \\
\hline
$\delta_{90\%}$ & 3.61 & 3.53 & 5.60 & 3.46 & \textbf{3.00} \\
\hline

\end{tabular}

\end{center}
\end{table}
\setlength{\tabcolsep}{1.4pt}

\section{Conclusions}

In this paper, we presented a novel approach composed of both deep learning and visual-geometry-based methods for estimating the complete 3D human head shape from a single or multiple images.
Our method employs an encode-decoder network that maps the input image to a rasterized normal map $\cal N$ and a landmark map $\cal Z$.
These maps are then used in a fitting process to regress the 3DMM parameters of the face identity from the LYHM model. To our knowledge, pixel-wise fitting is generally based on photometric information and not on output from a neural network, we strongly believe this process is intrinsically better than integration-based surface-reconstruction methods.

Our network has been trained only on a synthetic facial data set.
The network has shown good results in terms of both accuracy and generalization on real-world images.
In addition to the 3D reconstruction of the head, the landmark map $\cal Z$  can be directly used for face tracking and pose estimation, an essential part of our reconstruction pipeline. The landmarks are used to find a good initial guess for the pose before the fine-grained fitting. It improves the convergence rate of the regression process and minimizes chances of reaching a local minimum.

We have performed quantitative and qualitative experiments to evaluate our pipeline performance.
We demonstrate that our proposed framework achieves state-of-the-art performance in 3D face reconstruction for both single and multi-view settings.
Overall, the multi-view setting gives promising results but further work is required to take full advantage of it.

Despite the robust performance in many cases, our method has some limitations.
The used 3DMM does not include facial expressions and has a limited age range. This is why it is difficult to reconstruct the finest details in 3DMM because the accuracy of the recovered geometry is limited to the flexibility of this model.
This limitation is not fundamental to our proposed method. It can be overcome by adopting a more expressive morphable model for the synthetic data generator and for the fitting process.
Another limitation is that synthetic data can have unrealistic features, which in turn can introduce certain biases in the learning process.
The use of GAN architectures in combination with classic 3D rendering may improve the realism of generated photos.






%





\end{document}